\documentclass[twoside]{article}

\usepackage{PRIMEarxiv}

\usepackage[utf8]{inputenc} 
\usepackage[T1]{fontenc}    
\usepackage{hyperref}       
\usepackage{url}            
\usepackage{booktabs}       
\usepackage{amsfonts}       
\usepackage{nicefrac}       
\usepackage{microtype}      
\usepackage{lipsum}
\usepackage{fancyhdr}       
\usepackage{graphicx}       
\graphicspath{{media/}}     

\newcommand{\llm}{l}
\newtheorem{example}{Example}

\title{Leveraging Graph Structures and Large Language Models for End-to-End Synthetic Task-Oriented Dialogues}

\author{
  Maya Medjad \\
  UCBL, CNRS, Centrale Lyon, INSA Lyon,\\
  Univ. Lumière Lyon 2, LIRIS, UMR5205 \\
  69622 Villeurbanne, France\\
  \texttt{maya.medjad@univ-lyon1.fr} \\
   \And
  Hugo Imbert \\
  Reecall \\
  69000 Lyon, France \\
  \texttt{hugo@reecall.co} \\
  \AND
  Bruno Yun \\
  UCBL, CNRS, Centrale Lyon, INSA Lyon,\\
  Univ. Lumière Lyon 2, LIRIS, UMR5205 \\
  69622 Villeurbanne, France \\
  \texttt{bruno.yun@univ-lyon1.fr} \\
  \And
  Raphaël Szymocha \\
  Reecall \\
  69000 Lyon, France \\
  \texttt{raphael@reecall.co} \\
  \And
  Frédéric Armetta \\
  UCBL, CNRS, Centrale Lyon, INSA Lyon,\\
  Univ. Lumière Lyon 2, LIRIS, UMR5205 \\
  69622 Villeurbanne, France \\
  \texttt{frederic.armetta@univ-lyon1.fr} \\
}

\begin{document}
\maketitle

\begin{abstract}
Training task-oriented dialogue systems is both costly and time-consuming, due to the need for high-quality datasets encompassing diverse intents. Traditional methods depend on extensive human annotation, while recent advancements leverage large language models (LLMs) to generate synthetic data. However, these approaches often require custom prompts or code, limiting accessibility for non-technical users. We introduce GraphTOD, an end-to-end framework that simplifies the generation of task-oriented dialogues. Users can create dialogues by specifying transition graphs in JSON format. Our evaluation demonstrates that GraphTOD generates high-quality dialogues across various domains, significantly lowering the cost and complexity of dataset creation.
\end{abstract}

\section{Introduction}

\textit{Task-Oriented Dialogue Systems (TODS)} are increasingly used in domains like customer support, personal assistants, and enterprise solutions to help users achieve specific objectives through natural language conversations \cite{DBLP:conf/emnlp/BudzianowskiV19,DBLP:journals/ftir/GaoGL19,qin2023end,wang2022task}.

Traditional TODS rely on machine learning models trained on predefined schemas \cite{rastogi2020scalablemultidomainconversationalagents,DBLP:journals/ftir/GaoGL19}, but they struggle with complex/nuanced dialogues, especially when domain-specific data is scarce.
In contrast, LLM-based TODS enable more human-like and engaging responses \cite{jiang2023mistral,brown2020language,soudani2024surveyrecentadvancesconversational,DBLP:conf/emnlp/BudzianowskiV19}. However, these systems are prone to hallucinations \cite{hudeček2023llmsneedtaskorienteddialogue,ji2023survey}, needing fine-tuning for specific use cases \cite{liu2024toadtaskorientedautomaticdialogs,DBLP:journals/corr/abs-2404-14772}.

Fine-tuning LLM-based TODS requires large amounts of training data, with diverse and high-quality structures, which are costly and time-consuming to collect \cite{zhang2020recent}.
A diverse dataset of realistic dialogues is essential to allow these systems to grasp the nuances and unique patterns of human conversation.
This becomes even more problematic when several TODS are required, each for a different task, spanning different domains (e.g., hotel booking).

\begin{figure*}
    \centering
    \includegraphics[width=16cm]{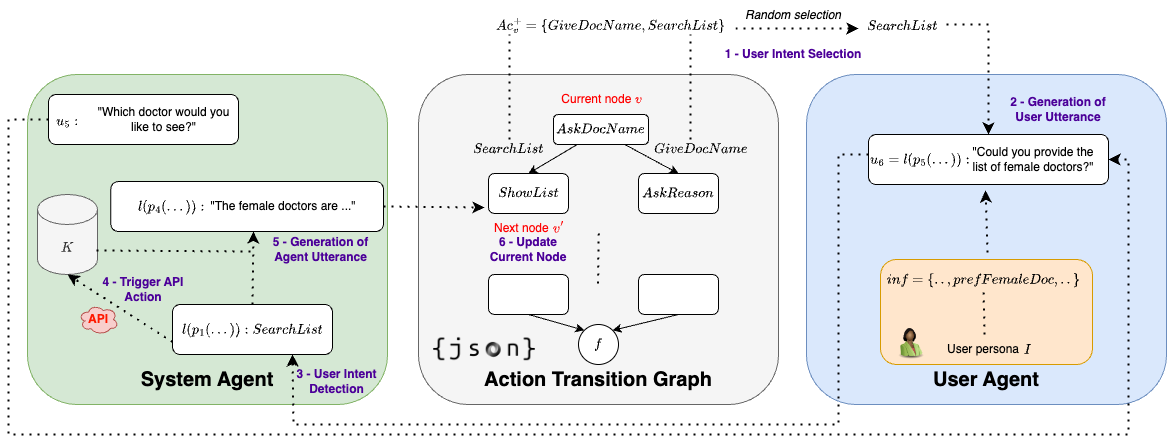}

    \caption{Representation of one turn of generation of the GraphTOD generation pipeline}
    \label{fig:graphTOD}

\end{figure*}

Data has previously been collected with human workers \cite{budzianowski2020multiwozlargescalemultidomain}, this type of dataset is particularly expensive to create even when users are assisted by computers \cite{rastogi2020scalablemultidomainconversationalagents,laps_joko}. 
Synthetic data generation is now a common approach to acquire data for TODS \cite{soudani2024surveyrecentadvancesconversational,stacey2024lucidllmgeneratedutterancescomplex}.

For example, \textit{SynTOD} \cite{DBLP:journals/corr/abs-2404-14772} proposes a new approach to model TODS behavior using a state transition graph. 
However, it was inaccessible to non-experts, as the graph and the corresponding prompts had to be implemented manually in Python. To alleviate this problem, we propose GraphTOD, a generalisable framework powered by LLMs with a new state machine-based prompt that allows non-technical users to generate task-oriented dialogues by specifying transition graphs in JSON format.
The source code and the demo video are available on Github\footnote{\url{https://github.com/reecall/GraphTOD}} and Youtube\footnote{\url{https://youtu.be/5pXa4yGcc58}} respectively.
\section{The GraphTOD generation pipeline}

GraphTOD is based on two agents (system and user) which simulate dialogue utterances by navigating an \textit{action transition graph} (see Figure \ref{fig:graphTOD}). We formalize each of those elements.

\textbf{An action transition graph} is a tuple $G = (V, Ac, E, t, s, f)$, where $V$ is a set of nodes, $Ac$ is a set of actions, $s,f \in V$ are the initial and final states, $E \subseteq (V \setminus \{f\}) \times Ac$ is a set of available actions at each non-final node, and $t: (V \setminus \{f\}) \times Ac \to V$ is the transition function.
%
The action transition graph serves as the link between the two agents and can be specified quickly in JSON format.

Given an action transition graph $G$, a subset of actions $F \subseteq Ac$, called \textit{function calls}, are associated with APIs to obtain external knowledge. For a node $v \in V$, the possible actions at $v$ are denoted $Ac_v^+ = \{ a \in Ac \mid (v,a) \in E\}$.
Our pipeline makes use of five carefully crafted prompt templates\footnote{We refer the reader to the Github repository for more details.} denoted by $p_i$, $1 \leq i \leq 5$. These prompt templates take as input a set of parameters and return a formatted string for an LLM.
A dialogue history is $H = (u_1, u_2, u_3, \dots, u_n)$, where $u_{2j}$ and $u_{2j+1}$ represent the user's and system's utterances, respectively, at time $j \geq 0$.
We denote an LLM as a (possibly non-deterministic) function $\llm$ that outputs $\llm(x)$ for input $x$.

\textbf{The system agent} is defined as $A_s= (G, F, P_s, K, u_1)$, where $G$ is an action transition graph, $F$ is a set of function calls, $P_s = \{ p_1, \dots, p_4\}$ is a set of system prompt templates, $K$ is an agent knowledge database initialized at $\emptyset$, and $u_1$ is a starting utterance. 
Considering a dialogue history $H= (u_1, u_2, \dots, u_{2j+1})$, a user utterance $u_{2j+2}$, and the current node $v \in V$, the system agent performs a two-steps reasoning.
First, $p_1(H, u_{2j+2}, K, Ac_v^+)$ is used to make the LLM detect the user's intention. Second, depending on the detected intention, potential APIs are triggered to collect knowledge, and the system utterance is generated to either state that the intent was not recognized, end the conversation, or continue the conversation (using the corresponding prompt templates $p_2$, $p_3$, or $p_4$).





\textbf{The user agent} is defined as $A_u = (I, P_u)$, with $P_u = \{ p_5\}$ a set of user prompt templates, and $I = ( age, name, gender, prefs)$ the agent's persona, where $age \in \{18, 19, \dots, 80\}, name$ is generated using the Faker library\footnote{\url{https://faker.readthedocs.io/}. $name$ is kept consistent with the agent's $gender$.}, $gender \in \{male, female\}$, and $prefs$ is a list of preferences generated based on $G$ and can be topics or places linked to it.
For $v \in V$ and $a \in Ac_v^+$, the user agent uses $p_5(a,G, H,I)$ to generate the next user utterance reflecting action $a$ at $v$.


\begin{example}
We illustrate the generation of a dialogue between a medical chatbot assistant and a user, guided by an action transition graph $G$ (see Figure \ref{fig:graphTOD}). At node $v= AskDocName$, the system agent initiates with the utterance $u_5 =$ ``Which doctor would you like to see?''. A random user intention $SearchList \in Ac_v^+$ is selected as the next action. Based on this intent and the user preference for female doctors ($prefFemaleDoc$), the user agent formulates the response $u_6 = $``Could you provide the list of female doctors?''. The system agent processes the user intent ($SearchList$), queries the relevant API to retrieve the list of doctors, and generates the next system utterance $u_7 =$``The female doctors are...''. Finally, the current node is updated to $ShowList$.

\end{example}

\section{Evaluation}

We generated 150 conversations on four domain scenarios (Recipe, Hotel, RentCar, Doctor) using our pipeline and OpenAI's GPT-4\footnote{Refer to the Github repository for the corresponding action transition graphs.}.
We evaluate the dialogues using three metrics (naturalness, coherence, understandability) from the pre-trained T5-based UniEval metrics \cite{zhong2022unifiedmultidimensionalevaluatortext} as classic NLP metrics (e.g., BLEU \cite{bleueval} or ROUGE \cite{lin2004rouge}) are not sufficient to portray the difference between the advanced generation models. 
As shown in Table \ref{table-perf-graphTOD}, GraphTOD performs consistently well overall and reports similar performances to human-in-the-loop approaches based on LLMs such as LAPS \cite{Joko_2024}.




\begin{small}
\begin{table}[h!]
\centering
\begin{tabular}{|l|c|c|c|c|c|}
\hline
\textbf{Model - Scenario}                                    & \textbf{Nat.} & \textbf{Coher.} & \textbf{Under.} & \textbf{Mean} \\ \hline
\hline
\textbf{(Baseline)} LAPS - Recipe & 0.867 & 0.891 & 0.860 & 0.872\\
\hline
\textbf{(Baseline)} LAPS - Movie & 0.874 & 0.897 & 0.868 & 0.880\\
\hline
\hline
\textbf{(Ours)} GraphTOD - Recipe                          & 0.899 & 0.857 & 0.890 & 0.882\\ \hline
\textbf{(Ours)} GraphTOD - Hotel                           & 0.888 & 0.853 & 0.879 & 0.873\\ \hline
\textbf{(Ours)} GraphTOD - RentCar                         & 0.887 & 0.768 & 0.878 & 0.844\\ \hline
\textbf{(Ours)} GraphTOD - Doctor                          & 0.835 & 0.800 & 0.827 & 0.820\\ \hline
\end{tabular}
\caption{Performance comparison with UniEval metrics.}
\label{table-perf-graphTOD}
\end{table}
\end{small}


\section{Conclusion}
GraphTOD is an end-to-end LLM-based pipeline designed to generate task-oriented conversations efficiently. By using an action transition graph in JSON format, GraphTOD simulates high-quality dialogues between two agents across various domains, thanks to a generalized prompting approach. GraphTOD also includes the automatic generation of user-agent preferences from the input graph and LLM-powered intent detection, resulting in a fully automated and fault-tolerant pipeline.

\newpage
\bibliographystyle{unsrt}  
\bibliography{sample}

\end{document}